\documentclass[3p,twocolumn]{elsarticle}

\usepackage{lineno,hyperref}
\usepackage{url}
\usepackage{hyperref}
\usepackage{graphicx}
\usepackage{subfigure}
\usepackage{booktabs}
\usepackage{threeparttable}
\usepackage{multirow}
\usepackage{url}
\usepackage{float}
\usepackage{amssymb}
\usepackage{amsmath}
\usepackage{amsmath}
\usepackage{bm}
\usepackage{blindtext}
\usepackage{mathrsfs}

\usepackage{color,xcolor}

\newcommand{\Revise}{\textcolor{black}}

\modulolinenumbers[5]

\begin{document}

\begin{frontmatter}

\title{Unconstrained Face Sketch Synthesis via Perception-Adaptive Network and A New Benchmark}

\author[scut]{Lin Nie}      
\author[polyu]{Lingbo Liu\corref{mycorrespondingauthor}} 
\author[sysu]{Zhengtao Wu}  
\author[scut]{Wenxiong Kang} 

\address[scut]{South China University of Technology, China}
\address[polyu]{Hong Kong Polytechnic University, Hong Kong}
\address[sysu]{Sun Yat-Sen University, China}

\cortext[mycorrespondingauthor]{Corresponding author}

\begin{abstract}
Face sketch generation has attracted much attention in the field of visual computing. However, existing methods either are limited to constrained conditions or heavily rely on various preprocessing steps to deal with in-the-wild cases. In this paper, we argue that accurately perceiving facial region and facial components is crucial for unconstrained sketch synthesis. To this end, we propose a novel Perception-Adaptive Network (PANet), which can generate high-quality face sketches under unconstrained conditions in an end-to-end scheme. Specifically, our PANet is composed of:
{\bf{i)}} a Fully Convolutional Encoder for hierarchical feature extraction,
{\bf{ii)}} a Face-Adaptive Perceiving Decoder for extracting potential facial region and handling face variations, and
{\bf{iii)}} a Component-Adaptive Perceiving Module for facial component aware feature representation learning.
To facilitate further researches of unconstrained face sketch synthesis, we introduce a new benchmark termed WildSketch, which contains 800 pairs of face photo-sketch with large variations in pose, expression, ethnic origin, background, and illumination. Extensive experiments demonstrate that the proposed method is capable of achieving state-of-the-art performance under both constrained and unconstrained conditions. Our source codes and the WildSketch benchmark would be resealed on the project page  {\color{blue}\url{http://lingboliu.com/unconstrained_face_sketch.html}}.
\end{abstract}


\end{frontmatter}


\section{Introduction}
Face sketch synthesis, which aims to automatically generate sketches from face photographs, is a crucial task in computer vision \cite{liu2005nonlinear,wang2008face,wang2017data,zhang2019neural,bi2021face}. Due to its broad applications in real-world scenarios, such as law enforcement \cite{zhang2011coupled} and digital entertainment \cite{zhang2017compositional,zhu2019face}, this problem recently has attracted widespread research interests in both academic and industry communities \cite{peng2020universal,lin2021toward,li2021face,zhu2021learning}.

\begin{figure}[t]
\begin{center}
 \includegraphics[width=1\columnwidth]{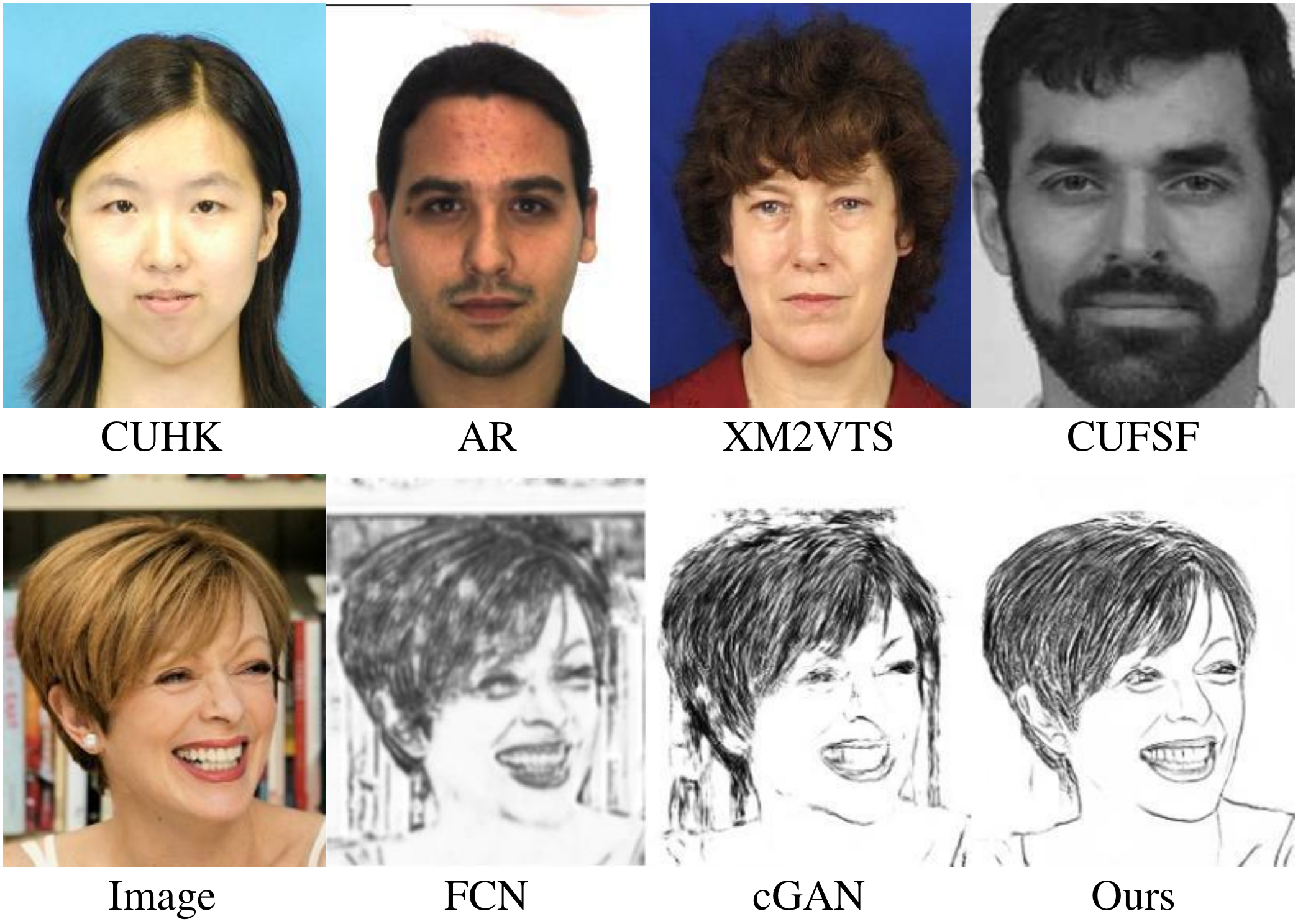}
 \vspace{-10mm}
\end{center}
   \caption{The first row shows the frontal faces with a pure background in four existing datasets. The second row visualizes an unconstrained facial image and the corresponding face sketches synthesized by FCN \cite{zhang2015end}, cGAN \cite{isola2017image} and our method. Note that FCN and cGAN have been finetuned on our WildSketch dataset.}
\vspace{0mm}
\label{fig:existing_methods}
\end{figure}

In the literature, numerous traditional models \cite{tang2003face,gao2008local,xiao2010photo,zhou2012markov} and deep learning models \cite{zhang2015end,isola2017image,zhang2018face,zhang2018markov,wan2021generative} have been proposed for face sketch synthesis. However, most previous methods were limited to synthesize face sketches under constrained conditions, i.e., frontal faces with a pure background. As shown in the second row of Figure \ref{fig:existing_methods}, these models become ineffective when the given faces have a large appearance change and the background is cluttered. Recently, some works \cite{zhang2015face,peng2015multiple,song2017fast} have attempted to deal with some slight facial variations, but their models may fail to handle serious variations. Despite certain progress, this problem is still far from being solved perfectly. Under this circumstance, more robust algorithms are desired to spark substantial progress in this field.

\begin{figure*}[t]
\begin{center}
 \includegraphics[width=2\columnwidth]{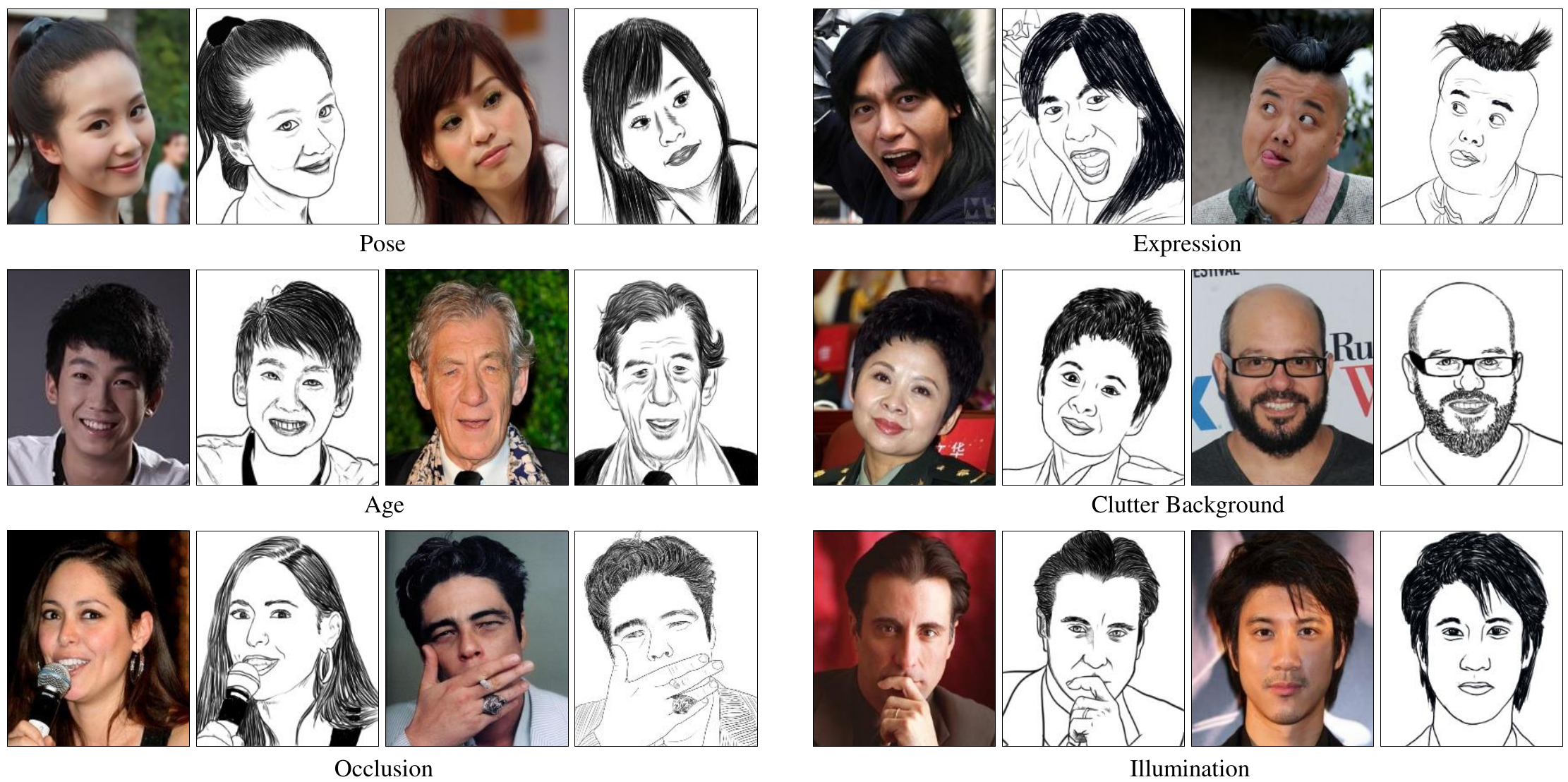}
 \vspace{-5mm}
\end{center}
   \caption{Samples from our WildSketch benchmark. The faces in our dataset are of huge variation in pose, expression, age and ethnic origin, while the background is complex and diverse.}
\vspace{-2mm}
\label{fig:sysu_sketch_example}
\end{figure*}

For unconstrained face sketch synthesis, it is crucial to distinguish the potential facial regions from cluttered background and handle the individual facial components (e.g., hair, eye, nose, and mouth) respectively \cite{zhang2016content,zhang2017compositional}. To this end, {\Revise{many approaches \cite{yi2019apdrawinggan,Li2021GENRE,yi2020line}}} heavily rely on various preprocessing, such as facial parsing \cite{smith2013exemplar,liu2020new} and facial landmark localization \cite{liu2019facial,deng2021geometry}. For instance, BFCN \cite{zhang2016content} needs to segment human faces from background in advance, while CA-GAN \cite{yu2020toward} has to decompose the given faces into different components before synthesizing their sketches. Nevertheless, these preprocessing procedures are time-consuming and may fail in unconstrained scenarios. This would severely degrade the performance of existing methods and weaken their flexibility in real-world applications.

In this paper, we propose a unified sketch synthesis framework termed Perception-Adaptive Network (PANet), which is able to generate high-quality face sketches under unconstrained conditions in an end-to-end scheme. Specifically, our PANet incorporates three tailor-designed components to perceive the potential facial region and each facial component adaptively.
First, a Fully Convolutional Encoder takes an unconstrained face image as input to perform hierarchical feature extraction. Second, a Face-Adaptive Perceiving Decoder learns the offsets of convolution dynamically to distinguish the face from the background and better capture the facial appearance variations. Third, after dividing the feature map of the unaligned face into multiple regions, a Component-Adaptive Perceiving Module is applied to dynamically compute the filter weights for the potential facial component in each local region based on its content. Finally, the feature of each region is convolved with its corresponding generated weights to optimally synthesize the sketch patch of the region.
It is worth noting that our method has two appealing properties. {\Revise{{\bf{i)}} Thanks to the face-component perception mechanism,
our synthesized face sketches can preserve more details and are more consistent with human perception, regardless of constrained or unconstrained settings. {\bf{ii)}} Compared with current methods \cite{yu2020toward,yi2020line,Li2021GENRE} for unconstrained scenes, our PANet has greater applicability and does not rely on any preprocessing techniques, e.g., face segmentation and alignment.}}

Moreover, all existing sketch synthesis datasets (e.g, CUHK student \cite{wang2008face}, AR \cite{MaB1998}, XM2VTS \cite{messer1999xm2vtsdb}, and CUFSF \cite{zhang2011coupled}) have a serious drawback in sample diversity. As shown in Figure \ref{fig:existing_methods}, these datasets only contain the frontal faces with a pure background. {\Revise{Although some works \cite{wan2021generative,yan2021isgan} have attempted to generate sketches under unconstrained conditions, there is no medium/large-scale dataset available for training and evaluation.}} To facilitate further researches in this field, we introduce a new benchmark termed ``WildSketch'', which focuses on sketch generation in the wild. Specifically, our WildSketch dataset has three appealing properties. First, it contains 800 pairs of face photo-sketch and it is four times larger than the popular CUHK and AR datasets. Second, the images in WildSketch are collected from real-world scenarios and these faces exhibit various poses, expressions, and age, and even have severe occlusions. Third, the background in our benchmark is more complex and diverse than all previous datasets.
In summary, the main contributions of our work are three-fold:
\begin{itemize}
\item We propose a novel Perception-Adaptive Network with three tailor-designed components, which can synthesize high-quality face sketches under unconstrained conditions without relying on any preprocessing.
\item We construct a new benchmark WildSketch, which contains 800 pairs of face photo-sketch with huge variations of appearance and background. To the best of our knowledge, WildSketch is the first {\Revise{medium-scale}} benchmark for unconstrained face sketch synthesis.
 \item Extensive experiments show that our proposed method achieves superior performance in comparison to existing state-of-the-art methods of face sketch synthesis under both constrained and unconstrained conditions.
\end{itemize}

The rest of this paper is organized as follows. We first review some related works of face sketch synthesis in Section \ref{sec:related works}. We then introduce the details of WildSketch dataset in Section \ref{sec:dataset} and describe the proposed method in Section \ref{sec:approach}. The performance of our approach is compared with numerous state-of-the-art methods in Section \ref{sec:experiment}. Finally, we conclude the paper in Section \ref{sec:conclusion}.

\section{Related Works}\label{sec:related works}
Human face analysis \cite{li2017face,liu2018facial,li2019semantic,chen2021cross,wang2020suppressing,yang2021larnet} is an important research topic in computer vision. As a classical face-related task, face sketch synthesis has been extensively studied. In literature, a large number of traditional approaches have been proposed and they can be divided into exemplar-based methods \cite{liu2007bayesian,zhou2012markov} and regression-based methods \cite{zhu2016simple}. In the past decade, deep neural networks (DNN) have achieved great success in various tasks \cite{chen2016disc,liu2019crowd,liu2019contextualized,chen2020knowledge,liu2020physical,liu2021cross}, and many researchers have also applied DNN for sketch synthesis \cite{lu2019fcn,bi2019multi,zhu2019deep,zhang2019deep,zhang2019bionic,zhu2020knowledge,yu2020toward}. For example, Zhang \textit{et al.} \cite{zhang2015end} developed an end-to-end fully convolutional network to model the mapping between photos and sketches. Philip \textit{et al.} \cite{philip2017face} explored the current conditional generative adversarial network framework and applied it to face sketch synthesis. Chen \textit{et al.} \cite{chen2018face} proposed a style transfer approach to introduce textures and shadings based on a pyramid column feature. {\Revise{Zhu \textit{et al.} \cite{zhu2021learning} combined generative Probabilistic Graphical Model and discriminative deep patch representation to jointly model the parameter distribution for deep patch representation and sketch patch reconstruction. Subsequently, Zhu \textit{et al.} \cite{zhusketch} also employed a transformer \cite{vaswani2017attention} to capture the global structural information of face sketches. Yan \textit{et al.} \cite{yan2021isgan} introduced an identity recognition loss to generate identity-preserving sketches, which was important for heterogeneous face recognition \cite{cao2018data,cao2018asymmetric,caomulti,liu2021iterative}.}} Nevertheless, without capturing the face variations, these methods are only feasible under constrained conditions and usually fail in real-world scenarios.

Recently, some works \cite{zhang2015face,peng2015multiple,zhu2017deep,zhang2019cascaded} have attempted to generate face sketches under some slightly unconstrained conditions. However, many of them heavily relied on preprocessing to deal with certain variations of faces. For instance, $p$GAN \cite{zhang2018robust} required to exclude background from images before feeding the clean faces into the network. BFCN \cite{zhang2016content} needed to parse the given image into semantic regions of the face, hair, and background before the synthesis, while CA-GAN \cite{yu2020toward} further decomposed the input photo into multiple fine-grained components (e.g. hair, nose, eye, and mouth). {\Revise{APDrawings++ \cite{yi2020line} also had to detect the bounding box of each component before feeding it into the corresponding local network.}} To extract the geometric feature of faces, CMSG \cite{zhang2017compositional} needs to localize 194 facial landmarks in advance.

 \begin{table}
\newcommand{\tabincell}[2]{\begin{tabular}{@{}#1@{}}#2\end{tabular}}
  \centering
  \caption{Overview of the publicly available datasets for face sketch synthesis.}
  \vspace{2mm}
  \resizebox{7.9cm}{!} {
  \begin{tabular}{c|c|c|c|c}
    \hline
    Dataset & Total  & Train & Test & Setting  \\ \hline
    CUHK        & 188    & 88    & 100  &  \multirow{4}{*}{constrained}  \\ \cline{1-4}
    AR          & 123    & 80    & 43   &    \\ \cline{1-4}
    XM2VTS      & 295    & 100   & 195  &    \\ \cline{1-4}
    CUFSF       & 1194   & 250   & 944  &    \\ \hline
    WildSketch  & 800    & 400   & 400  & unconstrained \\ \hline
  \end{tabular}
  }
  \vspace{0mm}
  \label{tab:dataset}
\end{table}

\begin{figure*}[t]
\begin{center}
 \includegraphics[width=2.1\columnwidth]{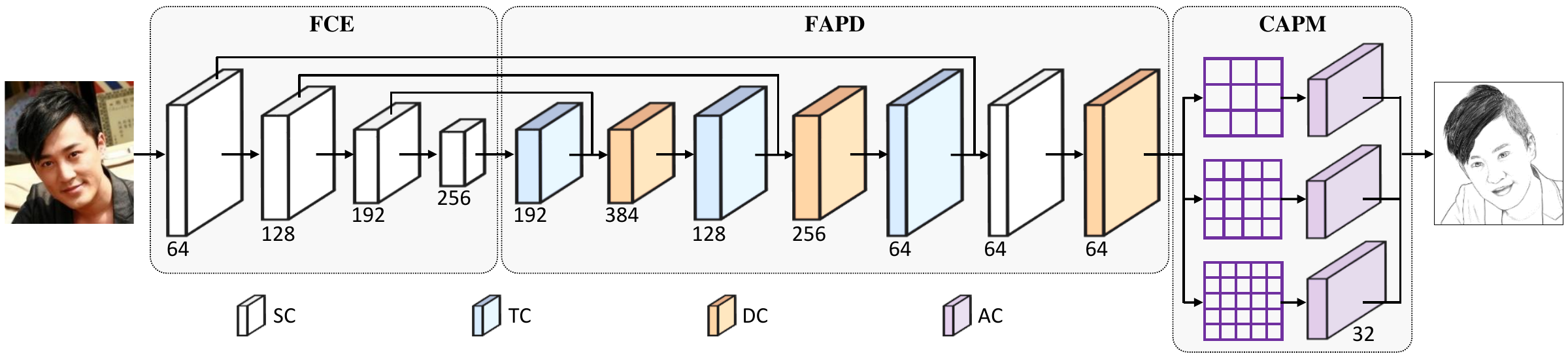}
 \vspace{-9mm}
\end{center}
   \caption{The architecture of the proposed Perception-Adaptive Network for unconstrained face sketch synthesis. Our network consists of a Fully Convolutional Encoder, a Face-Adaptive Perceiving Decoder (FAPD), and a Component-Adaptive Perceiving Module (CAPM). SC and TC denote a standard convolutional layer and transposed convolutional layer, respectively. DC and AC are a deformable convolutional layer and adaptive convolutional layer. A block with the text ``$c$'' denotes that its output feature has $c$ channels.}
\label{fig:network}
\vspace{0mm}
\end{figure*}

However, aforementioned preprocessing methods may fail in complex scenes, which would greatly restrict the flexibility of these models. In contrast, our method directly takes unconstrained images as input to synthesize face sketches by incorporating deformable convolutions and adaptive convolutions. {\Revise{Due to the face-component perception mechanism, our method not only can generate high-quality face sketches but also can be widely applied in unconstrained scenarios.}}

\section{WildSketch Dataset}\label{sec:dataset}
In this section, we introduce our WildSketch benchmark, the first dataset for unconstrained face sketch synthesis. We first collect face photographs from two different sources. Specifically, some are collected from the FaceScrub dataset~\cite{ng2014data}, which contains 530 identities and most of them are Western celebrities. For each identity, we randomly choose an image and put it into our benchmark. To enlarge the ethnicity coverage, we collect other 270 images of Asian celebrities from the Internet.
Then, we invite artists to draw a sketch of each image {\Revise{on an electronic drawing book, where the face image is displayed on the screen as a reference layer\footnote{In previous works \cite{wang2008face,MaB1998,messer1999xm2vtsdb,zhang2011coupled}, artists were requested to draw the sketches of human faces on paper, thus their sketch images and face images were usually unaligned.}. Finally, our dataset contains 800 pairs of highly-aligned face photo-sketch, which is of better quality and larger scale than the popular CUHK dataset and AR dataset.}} Furthermore, 400 pairs are randomly chosen for training and the rest are for testing. The overview of our benchmark and existing publicly available datasets are summarized in Table \ref{tab:dataset}. Our WildSketch is much more challenging, as there are more variations in pose, age, expression, background clutters, and illumination, as shown in Figure \ref{fig:sysu_sketch_example}. We do hope that the proposed WildSketch benchmark could significantly promote researches in this field.

\section{Method}\label{sec:approach}
As illustrated in Figure \ref{fig:network}, we propose a novel Perception-Adaptive Network (PANet) for face sketch synthesis in the wild. {\Revise{Specifically, our PANet is composed of {\bf{i)}} a Fully Convolutional Encoder for hierarchical feature
extraction, {\bf{ii)}} a Face-Adaptive Perceiving Decoder for extracting potential facial regions and handling face variations, and {\bf{iii)}} a Component-Adaptive Perceiving Module for facial component aware feature representation learning.}} First, the FCE takes an unconstrained image as input and extracts the hierarchical features with consecutive standard convolutional layers and pooling layers. Then, incorporating the deformable convolutional layers \cite{dai2017deformable}, the FAPD dynamically learns the offset of convolution and implicitly attends the potential face regions to better capture the facial appearance variation. Third, the CAPM divides the feature map of the unaligned face into multi-scale regions, each of which may contain a specific facial component. For each region, CAPM dynamically generates the filter weights based on its content with adaptive convolutional layers \cite{kang2017incorporating}, and then convolve its feature with the computed weights to generate a specialized feature for the potential facial component in this region. Finally, the filtered features of all regions are combined to synthesize a high-quality sketch.

\subsection{Fully Convolutional Encoder}
Given an image taken in an unconstrained scenario, previous methods~\cite{zhang2016content,zhang2018robust} typically need to parse the face into multiple semantic regions or remove background with external models. In contrast, we directly feed the image into our Fully Convolutional Encoder to extract the hierarchical features. Specifically, our FCE consists of eight standard convolutional layers, each of which is followed by a Relu layer.
For computational efficiency, we implement these convolutional layers with lightweight filters. The kernel size of these layers is $3\times3$ and their output channel numbers are 64, 64, 128, 128, 192, 192, 256, and 256 respectively.
Moreover, after the second, fourth, and sixth convolutional layers, a pooling layer is inserted to downsample the features and its stride is $2\times2$. Assume that the resolution of the input image is $H\times W$, then the dimension of the final output feature of FCE is $256 \times \frac{H}{8} \times \frac{W}{8}$.

\begin{figure}[t]
\begin{center}
 \includegraphics[width=0.85\columnwidth]{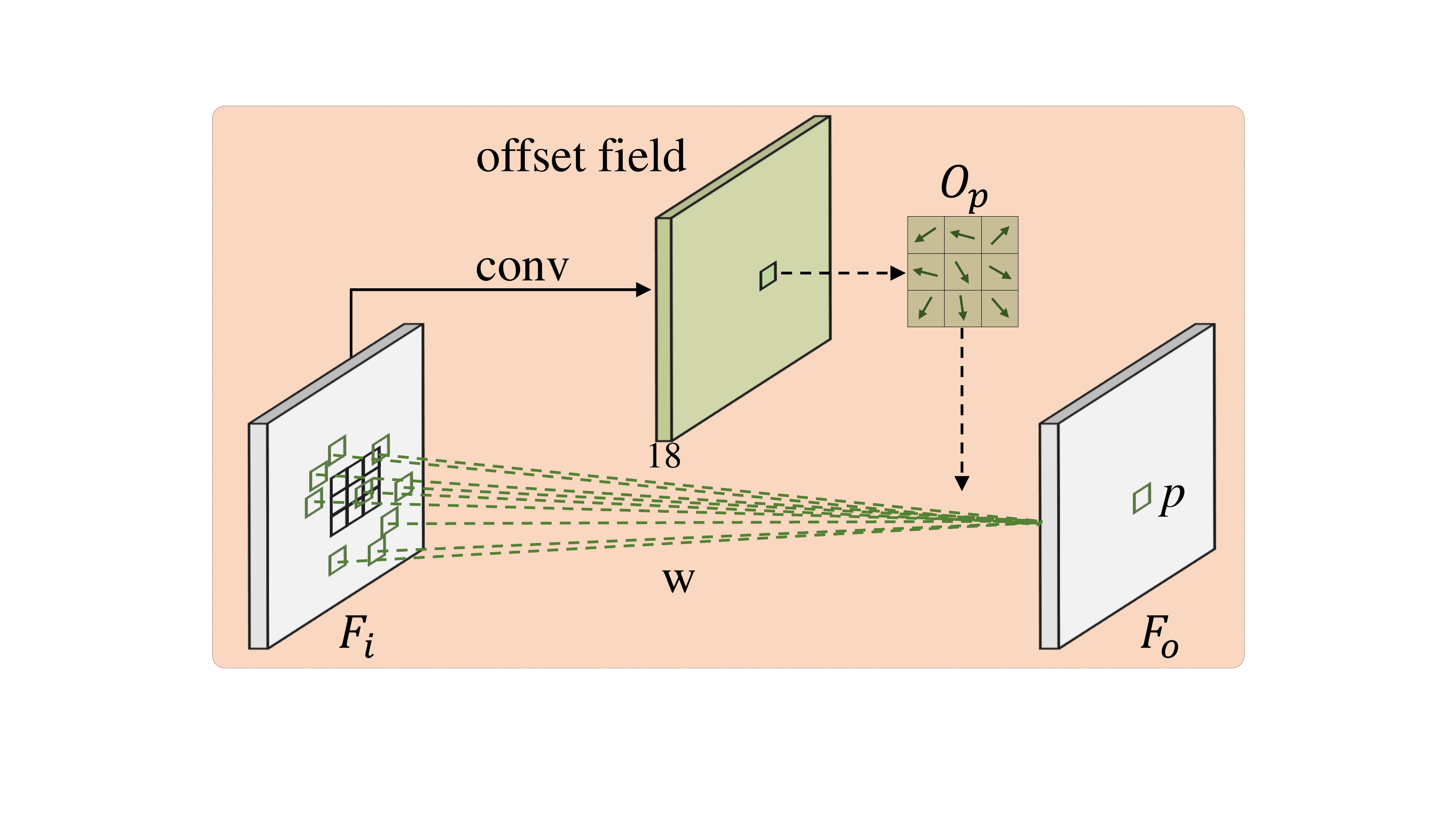}
 \vspace{-4.5mm}
\end{center}
   \caption{The architecture of the deformable convolutional layer implemented in our method. Its parameter is denoted as $w$ and its kernel size is $3\times 3$. $F_i$ and $F_o$ are its input and output features. $p$ is a location and the $3\times3$ green map $O_p$ is the offset of $p$.}
\vspace{0mm}
\label{fig:deformable_conv}
\end{figure}

\subsection{Face-Adaptive Perceiving Decoder}
In FAPD, we upsample the output features of FCE gradually with transposed convolutional layers and reconstruct the sketch image. During the decoding process, our FAPD suppresses the cluttered background and focuses on face modeling. To this end, our FAPD is designed to decodes the features with deformable convolutional layers, which dynamically learn the location offsets of convolution. With these offsets, the deformable convolution can generalize various transformations for pose scale, aspect ratio, and rotation, which can better fit the appearance of human faces in the wild. Moreover, we also find that the offsets of a foreground pixel can well cover the whole face (See Figure \ref{fig:deform_visual}) and help our network to better capture the face variations.

The architecture of the deformable convolutional layer implemented in our FAPD is shown in Figure~\ref{fig:deformable_conv}. We denote its input and output features as $F_i$ and $F_o$ respectively, and assume they have the same dimension.
Specifically, the deformable convolutional layer contains two steps. {\bf{i) Generate Offset:}} we feed $F_i$ into a standard convolutional layer with a kernel size of $3\times3$ to dynamically compute an offset field $O$ with 18 channels for all pixels. For a pixel $p$, its offsets can be organized as a $3\times3$ offset map $O_p$.
{\bf{ii) Conduct Convolution:}} we convolve the feature $F_i$ with the filter parameter $w$ of deformable convolutional layer under the guidance of the offset field $O$ to generate the output feature ${F}_o$. We first define a $3\times3$ grid $\mathcal{G}$ as \[
\mathcal{G}=\{(-1, -1), (-1, 0), \ldots, (0,1), (1, 1)\}
\]
For the location $p$ on the output feature map $F_o$, we have
\begin{equation}
F_o(p)=\sum_{G_k\in\mathcal{G}}w(G_k)\cdot F_i(p+G_k+O_p(G_k)),
\label{eq.deformable_conv}
\end{equation}
where $G_k$ enumerates the locations in $\mathcal{G}$.

As shown in Figure \ref{fig:network} (middle), our decoder consists of three deformable convolutional layers and three transposed convolutional layers with an upsampling rate of 2. After each transposed convolutional layer, we concatenate the upsampled feature and an FCE feature with the same resolution via skip connection to maintain the detail of faces. After the third skip connection, we apply a standard convolutional layer on top of the concatenated feature to reduce its channel number from 128 to 64 and then put the output feature into the third deformable convolutional layer.

\subsection{Component-Adaptive Perceiving Module}
In this subsection, we aim to handle each facial component individually. {\Revise{However}}, it is hard to obtain accurate facial components in complex scenarios. {\Revise{To this end, we propose a Component-Adaptive Perceiving Module to model each potential facial component effectively. Specifically, we first}} divide the feature map of the face image into multiple regions, each of which may contain a facial component. Incorporating pyramidal adaptive convolution layers, we dynamically generates the filter weights for each region based on its content. Finally, we convolve the features of each region with its individual filters to optimally generate the sketch of the region.

\begin{figure}[t]
\begin{center}
 \includegraphics[width=0.9\columnwidth]{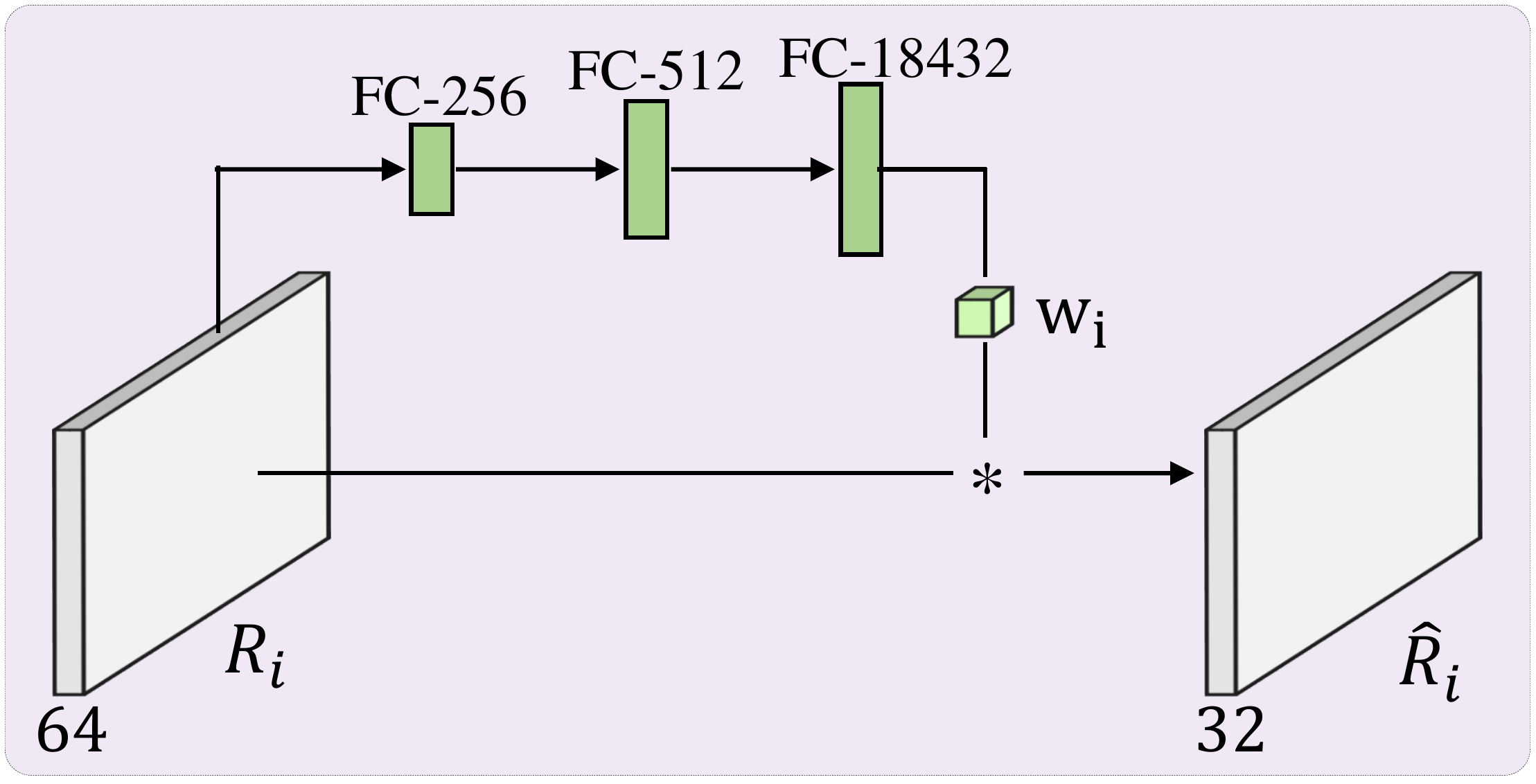}
 \vspace{-5mm}
\end{center}
   \caption{The architecture of adaptive convolutional layer implemented in our method. $R_i$ is the feature with 64 channels of the $i^{th}$ region. $w_i \in R^{64\times3\times3\times32}$ is the generated filters weight for $R_i$. $\hat{R}_i$ is the filtered feature with 32 channels. FC-$N$ is a fully connected layer with $N$ output neurons. ``*'' denotes a convolution operation. In our work, grouped FC layers are adopted for reducing the number of parameters.}
\label{fig:adaptive_conv}
\vspace{0mm}
\end{figure}

{\Revise{The details of our CAPM are described below.}} We first partition the output feature of FAPD into $n\times n$ regions and the resolution of each region is $\frac{H}{n} \times \frac{W}{n}$. The feature of $i^{th}$ region is denoted as $R_i$.
We then feed $R_i$ into an adaptive convolutional layer, which is shown in Figure \ref{fig:adaptive_conv}. Specifically, the adaptive convolutional layer contains two steps.
{\bf{i) Generate Parameter:}} we first use a $32\times32$ bins spatial pooling layer~\cite{he2015spatial} to transform $R_i$ into a fixed-size feature with low dimensionality, which facilitates reduction of the number of parameters in our network. We then feed the pooled feature into three grouped fully-connected (FC) layers with 256, 512, and 18,432 neurons respectively. The output of the last FC layer is reshaped to a filters weight $w_i \in R^{64\times3\times3\times32}$. The weight $w_i$ is specialized for the potential facial component in the $i^{th}$ region.
{\bf{ii) Conduct Convolution:}} we then convolve the feature $R_i$ with the generated filters $w_i$ and generate a new specialized feature $\hat{R}_i$ with 32 channels. {\Revise{Finally, the output features of $n\times n$ regions are reorganized to a feature with a resolution of $H \times W$.}}

{\Revise{However, we notice that simply}} dividing the feature map of FAPD into single-scale regions can not perfectly cover the facial components. To alleviate this issue, we partition the feature into multi-scale regions. As shown in the right of Figure \ref{fig:network}, our CAPM consists of three branches. Specifically, the top branch divides the feature map into $3\times3$ regions, while the middle and the bottom branches divide it into $4\times4$ regions and $5\times5$ regions, respectively. We also explore the different settings of CAPM in the experiment. The features of these multi-scale regions are fed into the adaptive convolutional layer to obtain the specialized features.
{\Revise{Finally, we concatenate the features of all branches and feed them into a convolutional layer with a kernel size of $1\times1$ to synthesize the final face sketch.}}

\begin{figure}[t]
\begin{center}
 \includegraphics[width=1\columnwidth]{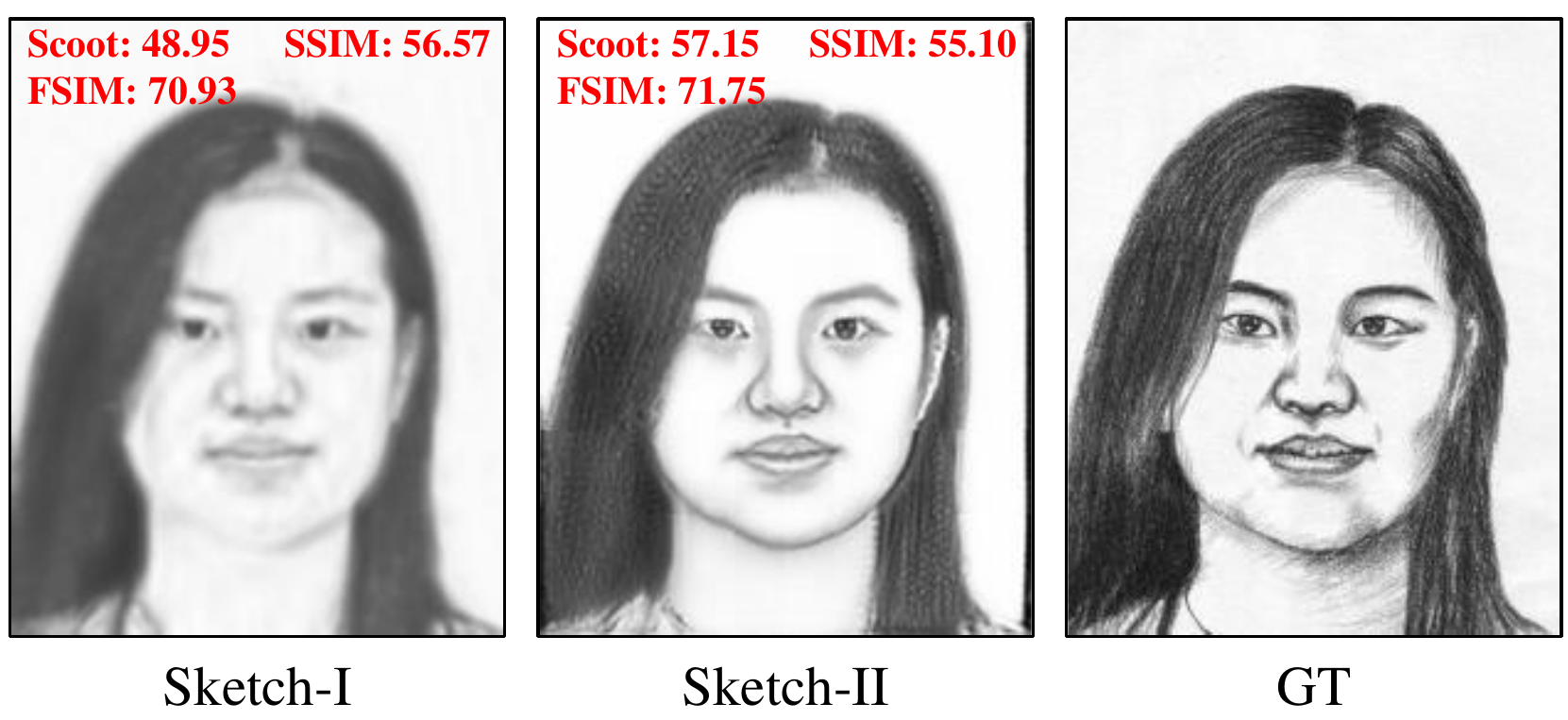}
 \vspace{-10mm}
\end{center}
   \caption{The illustration of the inconsistency between SSIM and human perception. In terms of perception, the generated sketch-II is much better than sketch-I. However, the SSIM score of sketch-II is much worse than that of sketch-I, since SSIM favors slightly blurry images. Therefore, the SSIM metric isn't adopted in this work.}
\vspace{0mm}
\label{fig:SSIM}
\end{figure}

\subsection{Implementation Detail}
We implement the proposed method with the Pytorch toolbox. The filter weights of all convolutional layers and fully connected layers are initialized by a normal distribution with a deviation equal to 0.02. The learning rate is set to 2e-4 and the batch size is 1. Similar to \cite{zhang2018robust}, we train our networks with the combination of Euclidean loss and adversarial loss. Our discriminator is the same as \cite{zhu2017unpaired}. Finally, we adopt the Adam~\cite{kingma2014adam} to optimize the whole network.

\section{Experiment}\label{sec:experiment}
In this section, we first introduce the evaluation metrics for face sketch synthesis. We then compare the proposed method with state-of-the-art methods under both the constrained and unconstrained conditions. Finally, we conduct extensive ablation studies to demonstrate the effectiveness of each component of our model.

\begin{table}[t]
\newcommand{\tabincell}[2]{\begin{tabular}{@{}#1@{}}#2\end{tabular}}
  \centering
  \caption{The performance of different methods {\bf{on the unconstrained WildSketch dataset}}. Our method is better than or comparable to CA-GAN and GENRE, both of which relied on preprocessing techniques. The results of the top two performance are highlighted in {\textcolor{red}{red}} and {\textcolor{blue}{blue}}, respectively.}
  \vspace{2mm}
    \begin{tabular}{c|c|c|c}
    \hline
     \tabincell{c}{Method}  & \tabincell{c}{Scoot$\uparrow$} & \tabincell{c}{FSIM$\uparrow$}  & \tabincell{c}{FID$\downarrow$} \\
    \hline\hline
     LLE        & 26.00 & 59.99 & 109.90 \\
     Fast-RSLCR & 24.53 & 59.99 & 115.08 \\
     RSLCR      & 24.33 & 60.03 & 106.95 \\
     FCN        & 25.59 & 58.27 & 104.06 \\
     BP-GAN     & 25.57 & 58.84 & 130.50 \\
     SSL        & 25.19 & 63.75 & 106.74 \\
     MDAL       & 31.51 & 68.79 & 49.42 \\
     CycleGAN   & 34.84 & 66.54 & {\bf\textcolor{blue}{45.94}} \\
     cGAN       & 35.11 & 67.50 & 66.74 \\
     GENRE      & 36.40 & 69.02 & 48.05 \\
     CA-GAN     & {\bf\textcolor{red}{37.40}} & {\bf\textcolor{red}{69.60}} & 49.38 \\
     Ours       & {\bf\textcolor{blue}{37.28}} & {\bf\textcolor{blue}{69.50}} & {\bf\textcolor{red}{38.73}} \\
    \hline
    \end{tabular}
  \vspace{0mm}
  \label{tab:Unconstrained_Result}
\end{table}

\subsection{Evaluation Metric}
In previous works, SSIM \cite{wang2004image} was widely used to measure the similarities between the generated face sketches and the ground-truth (GT) sketches. However, we observe that SSIM favors slightly blurry images and it could be inconsistent with the human perception of face sketches \cite{chen2018semi,wang2018back}, as shown in Figure \ref{fig:SSIM}. In this paper, we evaluate the quality of synthesized sketches with {\Revise{three}} more reasonable metrics, i.e., Scoot \cite{fan2018face}, FSIM \cite{zhang2011fsim} and {\Revise{FID \cite{heusel2017gans}}. Specifically, Scoot is a robust style similarity measure recently proposed for face sketch synthesis. {\Revise{It evaluates the block-level spatial structure and co-occurrence texture statistics simultaneously to provide consistent scores with human perception. FSIM measures the image similarity using the phase congruency and the gradient magnitude, and it is closely related to the subjective evaluation of humans \cite{wang2014heterogeneous}. FID utilizes the deep representation extracted by the Inception v3 model \cite{szegedy2016rethinking} to measure the Fréchet distance between two sets of images.}}

\begin{figure*}
\begin{center}
 \includegraphics[width=2\columnwidth]{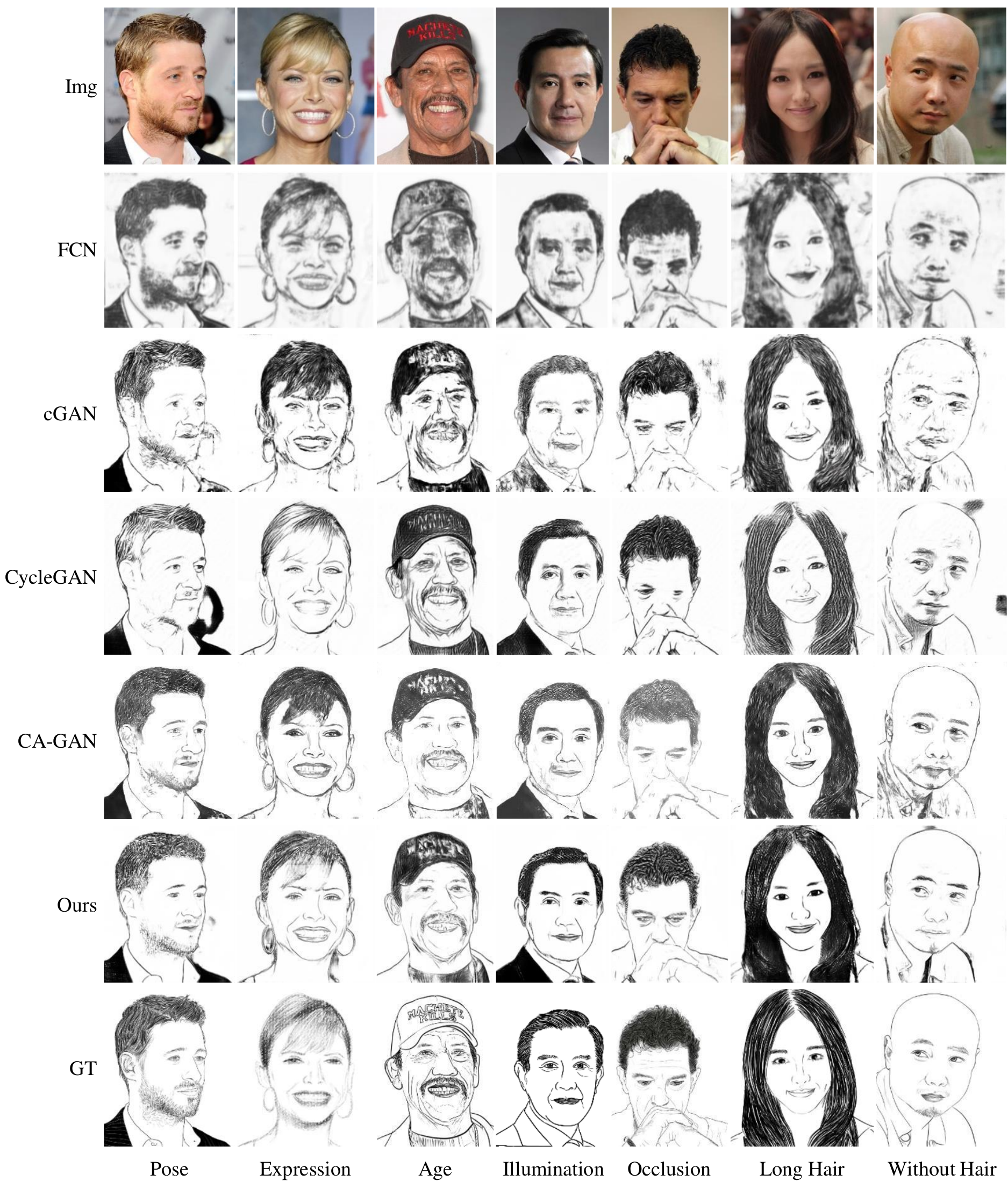}
 \vspace{-6mm}
\end{center}
   \caption{Visualization of the synthesized face sketches of different methods under unconstrained conditions. {\Revise{Without relying on any preprocessing techniques}}, our PANet can generate high-quality sketches with rich structure and texture details, even though these faces have huge variations of pose, expression, age, illumination, and cluttered background. {\Revise{More quantitative results can be found on our project page.}}}
\vspace{0mm}
\label{fig:Unconstrained_Result}
\end{figure*}

\begin{table*}[t]
\newcommand{\tabincell}[2]{\begin{tabular}{@{}#1@{}}#2\end{tabular}}
  \centering
  \caption{The performance of different methods {\bf{on four constrained face sketch datasets}}. The proposed method achieves the best results on most evaluation metrics. The results of the top two performance are highlighted in {\textcolor{red}{red}} and {\textcolor{blue}{blue}}, respectively.}
  \vspace{0mm}
  \resizebox{16.8cm}{!} {
    \begin{tabular}{c|c|c|c|c|c|c|c|c|c|c|c|c}
    \hline
    \multirow{2}{*}{Method}  & \multicolumn{3}{c|}{CUHK} & \multicolumn{3}{c|}{AR} & \multicolumn{3}{c|}{XM2VTS}& \multicolumn{3}{c}{CUFSF} \\
    \cline{2-13}
    & \tabincell{c}{Scoot$\uparrow$}  & \tabincell{c}{FSIM$\uparrow$} & \tabincell{c}{{\Revise{FID$\downarrow$}}}
    & \tabincell{c}{Scoot$\uparrow$}  & \tabincell{c}{FSIM$\uparrow$} & \tabincell{c}{{\Revise{FID$\downarrow$}}}
    & \tabincell{c}{Scoot$\uparrow$}  & \tabincell{c}{FSIM$\uparrow$} & \tabincell{c}{{\Revise{FID$\downarrow$}}}
    & \tabincell{c}{Scoot$\uparrow$}  & \tabincell{c}{FSIM$\uparrow$} & \tabincell{c}{{\Revise{FID$\downarrow$}}} \\
    \hline\hline
    FCN      & 46.20 & 70.75 & {\Revise{103.37}} & 53.79 & 72.87 & {\Revise{99.95}} & 42.90 & 67.86 & {\Revise{93.40}} & 43.78 & 66.24 & {\Revise{114.03}} \\
    SSD      & 46.80 & 71.77 & {\Revise{103.40}} & 55.21 & 74.86 & {\Revise{104.11}} & 42.57 & 67.29 & {\Revise{82.00}} & 46.87 & 68.23 & {\Revise{66.27}}\\
    RSLCR    & 46.80 & 72.28 & {\Revise{100.17}} & 54.29 & 74.37 & {\Revise{103.14}} & 42.00 & 67.27 & {\Revise{81.54}} & 45.31 & 66.49 & {\Revise{73.16}} \\
    Bayesian & 47.60 & 72.44 & {\Revise{79.70}}  & 55.95 & 74.86 & {\Revise{83.33}} & 44.22 & 68.31 & {\Revise{59.87}} & 45.92 & 67.54 & {\Revise{60.71}} \\
    DGFL     & 47.27 & 73.32 & {\Revise{88.24}}  & 55.88 & 75.44 & {\Revise{86.58}} & 43.32 & 68.46 & {\Revise{61.01}} & 47.24 & 69.56 & {\Revise{52.49}} \\
    MWF      & 48.86 & 73.35 & {\Revise{85.87}}  & 57.87 & 75.99 & {\Revise{84.95}} & 45.93 & 69.46 & {\Revise{60.11}} & 48.81 & 70.29 & {\Revise{53.94}} \\
    BP-GAN   & 48.28 & 71.76 & {\Revise{93.79}}  & 57.26 & 74.08 & {\Revise{97.81}} & 43.72 & 66.44 & {\Revise{65.84}} & 49.35 & 68.14 & {\Revise{52.04}}\\
    MRF      & 50.57 & 72.44 & {\Revise{77.34}}  & 62.62 & 74.97 & {\Revise{75.37}} & 49.50 & 68.44 & {\Revise{71.39}} & 61.66 & 69.56 & {\Revise{63.62}} \\
    RL+Dict  & 50.88 & 73.12 & {\Revise{73.51}}  & 60.26 & 74.78 & {\Revise{81.79}} & 47.93 & 70.92 & {\Revise{49.15}} & 48.37 & 69.44 & {\Revise{51.17}} \\
    SSL      & 51.34 & {\bf\textcolor{red}{74.31}} & {\Revise{80.88}}  & 58.93 & 76.77 & {\Revise{72.52}} & 45.83 & 71.05 & {\Revise{55.77}} & 50.84 & 71.02 & {\Revise{41.34}} \\
    CycleGAN & 55.10 & 72.49 & {\Revise{81.44}}  & 64.37 & 74.06 & {\Revise{81.95}} & 47.93 & 65.72 & {\Revise{66.02}} & 62.66 & 70.23 & {\Revise{26.03}} \\
    cGAN     & 53.33 & 72.96 & {\Revise{92.64}}  & 66.71 & 75.49 & {\Revise{74.99}} & 53.39 & 69.93 & {\Revise{55.52}} & 62.58 & 70.59 & {\Revise{27.13}} \\
    MDAL     & 55.08 & {\bf\textcolor{blue}{74.12}} & {\Revise{82.38}}  & 68.13 & {\bf\textcolor{blue}{76.71}} & {\bf\textcolor{red}{57.89}} & 52.72 & 71.17 & {\bf\textcolor{blue}{46.34}} & 62.60 & 70.76 & {\Revise{36.36}} \\
    GENRE    & {\bf\textcolor{red}{56.42}} & 72.27 & {\bf\textcolor{blue}{72.66}}  & {\bf\textcolor{blue}{69.32}} & 74.84 & 74.58 & 56.91 & 71.20 & {\bf\textcolor{red}{42.70}} & 63.70 & 71.77 & 36.52\\
    CA-GAN   & 55.93 & 73.88 & 78.51  & 69.18 & {\bf\textcolor{red}{77.30}} & 78.85 & {\bf\textcolor{blue}{57.22}} & {\bf\textcolor{red}{73.08}} & 50.33 & {\bf\textcolor{blue}{66.38}} & {\bf\textcolor{red}{72.68}} & {\bf\textcolor{blue}{29.28}}\\
    Ours     & {\bf\textcolor{blue}{56.22}} & 72.93 & {\bf\textcolor{red}{67.17}}  & {\bf\textcolor{red}{69.44}} & 76.53 & {\bf\textcolor{blue}{68.30}} & {\bf\textcolor{red}{57.81}} & {\bf\textcolor{blue}{72.89}} & 47.09 & {\bf\textcolor{red}{66.84}} & {\bf\textcolor{blue}{72.20}} & {\bf\textcolor{red}{24.19}} \\
    \hline
    \end{tabular}
  }
  \vspace{0mm}
  \label{tab:Constrained_Result}
\end{table*}

\subsection{Performance on Unconstrained Setting}
In this section, we compare our PANet with {\Revise{eleven}} state-of-the-art methods in uncontrolled scenarios, including LLE \cite{roweis2000nonlinear}, Fast-RSLCR \cite{wang2018random}, RSLCR \cite{wang2018random}, FCN \cite{zhang2015end}, BP-GAN \cite{wang2018back}, SSL \cite{chen2018semi}, MDAL \cite{zhang2018face}, CycleGAN \cite{zhu2017unpaired}, cGAN \cite{isola2017image}, {\Revise{GENRE \cite{Li2021GENRE}, and CA-GAN \cite{yu2020toward}}}. These methods are reimplemented with their official codes and are optimized on the WildSketch benchmark. {\Revise{Notice that both GENRE and CA-GAN require the parsing map of the given face image for sketch synthesis.}}

The quantitative results of all methods are summarized in Table \ref{tab:Unconstrained_Result} and our PANet achieves a Scoot of 37.28\% and an FSIM of 69.50\%, outperforming all other compared methods {\Revise{that also don't use preprocessing techniques. Compared with GENRE and CA-GAN, our method can also obtain very competitive performance, especially on the FID indicator.}}
Moreover, we visualize the synthesized sketches of our PANet and other two classical deep models in Figure \ref{fig:Unconstrained_Result}. We can observe that the synthesized sketches of FCN and cGAN are blurry and noisy. In contrast, due to the great capability of the Face-Adaptive Perceiving Decoder and Component-Adaptive Perceiving Module, our PANet is able to generate high-quality face sketches with rich structure and texture details, even though those faces contain huge variations of pose, expression, age, illumination, and cluttered background.

\subsection{Performance on Constrained Setting}
In this subsection, we apply the proposed PANet to generate face sketches under constrained conditions. We conduct  extensive experiments on four commonly used datasets, including the CUHK student dataset, AR dataset, XM2VTS dataset, and CUFSF dataset. The images in these datasets are frontal faces with a pure background. We follow previous works \cite{wang2018random,wang2018back} to split the training set and testing set. Table \ref{tab:dataset} shows the number of training samples and testing samples on each dataset.

We perform extensive comparisons against {\Revise{15}} state-of-the-art methods on these datasets, including FCN \cite{zhang2015end}, SSD \cite{song2014real}, RSLCR \cite{wang2018random}, Bayesian \cite{wang2017bayesian}, DGFL \cite{zhu2017deep}, MWF \cite{zhou2012markov}, BP-GAN \cite{wang2018back}, MRF \cite{wang2008face}, RL+Dict \cite{jiang2019graph}, SSL \cite{chen2018semi}, CycleGAN \cite{zhu2017unpaired}, cGAN \cite{isola2017image},  MDAL \cite{zhang2018face}, {\Revise{GENRE \cite{Li2021GENRE}, and CA-GAN \cite{yu2020toward}. Note that \cite{Li2021GENRE,yu2020toward} are reimplemented with their official codes.}} The synthesized sketches of the remaining methods are collected from RSLCR\footnote{http://www.ihitworld.com/RSLCR.html}, BP-GAN\footnote{http://www.ihitworld.com/WNN/Back\_Projection.zip} or receive them from the official authors. We measure the Scoot,  FSIM score and FID of all compared methods in Table\ref{tab:Constrained_Result}.
We can observe that the proposed PANet achieves superior performance in comparison to almost all compared methods. More importantly,
{\Revise{our method also outperforms the state-of-the-art models CA-GAN and GENRE on most evaluation metrics.}} Furthermore, we visualize the synthesized sketches of four deep learning-based methods in Figure \ref{fig:Constrained_Result}. We can observe that our PANet is still better than the existing methods, especially in some local regions (e.g., jaw and mouth), although they have achieved promising results.

\begin{table}[t]
\newcommand{\tabincell}[2]{\begin{tabular}{@{}#1@{}}#2\end{tabular}}
  \centering
  \caption{Scoot (\%) of different variants of the proposed PANet that is composed of FCE, FAPD, and CAPM. In this table, FAPD-SC refers to a variant of FAPD, which replaces the deformable convolutions with standard convolutions (SC).}
  \vspace{3mm}
    \begin{tabular}{c|c|c}
    \hline
    \tabincell{c}{Method} & \tabincell{c}{WildSketch} & \tabincell{c}{CUFSF}  \\
    \hline\hline
     FCE+FAPD-SC{~~~~~~~}  & 34.48 & 63.07\\
     \hline
     FCE+FAPD{~~~~~~~~~~~~} & 35.74 & 63.63 \\
     \hline
     FCE+FAPD+CAPM & 37.28 & 66.84\\
    \hline
    \end{tabular}
  \vspace{0mm}
  \label{tab:ablation}
\end{table}

\begin{figure}[t]
\begin{center}
 \includegraphics[width=1\columnwidth]{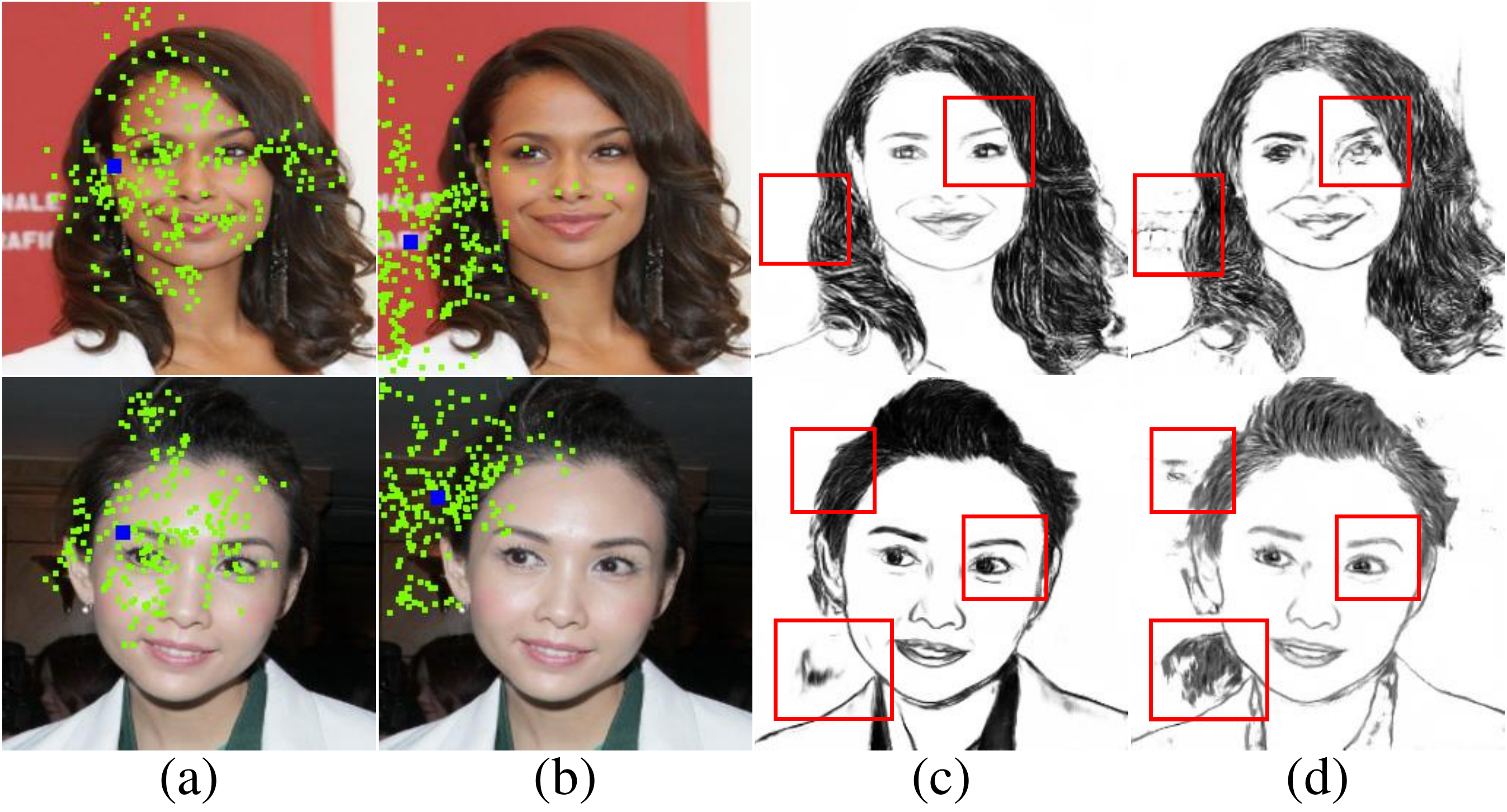}
 \vspace{-9mm}
\end{center}
   \caption{(a) and (b) show the learned $9^3$=729 offsets (green points) in FAPD for the activation units (blue points) on the face and background. (c) and (d) are the synthesized sketches of model ``FCE+FAPD" and ``FCE+FAPD-SC", respectively. (Better to zoom in.)}
\vspace{0mm}
\label{fig:deform_visual}
\end{figure}

\begin{figure*}[t]
\begin{center}
 \includegraphics[width=2\columnwidth]{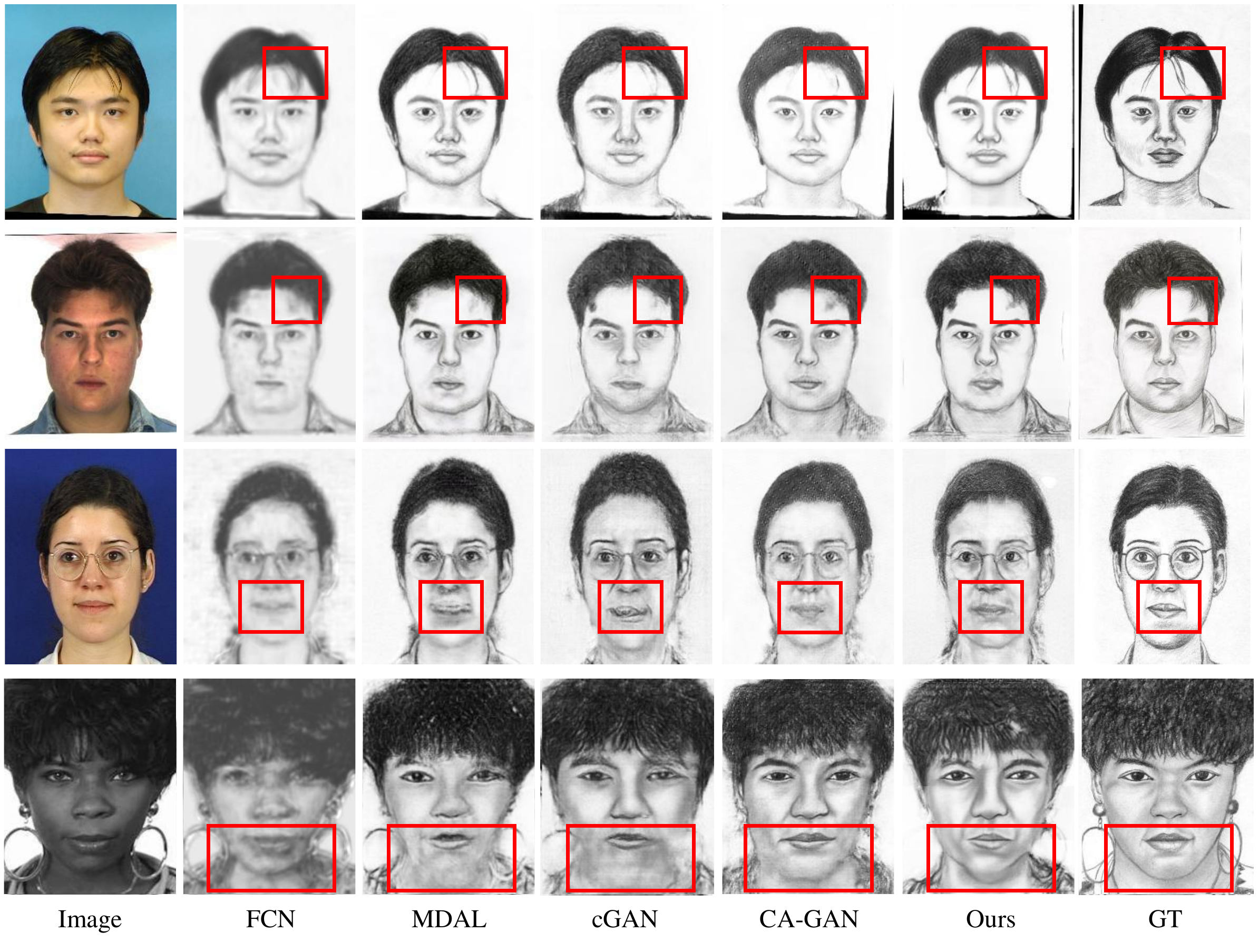}
 \vspace{-5mm}
\end{center}
   \caption{\Revise{Visualization of the synthesized face sketches of different methods under constrained conditions. These four rows show the results on the CUHK, AR, XM2VTS, and CUFSF, respectively. Compared with previous methods, our PANet can generate better face sketches, especially in some local regions. More quantitative results can be found on our project page.}}
\vspace{0mm}
\label{fig:Constrained_Result}
\end{figure*}

\subsection{Ablation Studies}
In this subsection, we perform extensive ablation studies on the unconstrained WildSketch dataset and the largest constrained CUFSF dataset.

\subsubsection{Effectiveness of Face-Adaptive Perceiving Decoder}
To verify the effectiveness of FAPD, we implement a variant FAPD-SC that removes the adaptive perception by replacing the deformable convolutions in FAPD with standard convolutions. As shown in Table \ref{tab:ablation}, compared with  ``FCE+FAPD-SC'', the model ``FCE+FAPD'' achieves an improvement of Scoot 1.26\% on WildSketch dataset and 0.56\% on the CUFSF datasets.
Meanwhile, we visualize the learned offset positions of FAPD in Figure \ref{fig:deform_visual}.
For a pixel in the face region, its 729 offsets\footnote{FAPD contains three deformable convolutional layers with a $3\times3$ kernel size (9 offsets), thus each pixel has $9^3$ = 729 offsets finally and some offsets may be overlapped.} expand to the whole face and only a few offsets are out of the face, which can guide our network to focus on the face region and better handle the variation of the face. For a pixel in the background, it enlarges the receptive field with offset and receives more content from different regions, which can facilitate our network to distinguish the background point.
Moreover, the synthesized sketches of model ``FCE+FAPD" and ``FCE+FAPD-SC" are shown in Figure \ref{fig:deform_visual}(c) and \ref{fig:deform_visual}(d), respectively. We can observe that the sketch of the former is much better than that of the latter, especially on some facial components and cluttered background.
Therefore, these quantitative and qualitative experiments have clearly demonstrated the effectiveness of FAPD.

\begin{table}[t]
\newcommand{\tabincell}[2]{\begin{tabular}{@{}#1@{}}#2\end{tabular}}
  \centering
  \caption{Scoot (\%) of different variants of CAPM on WildSketch dataset. The setting of ``3,4,5" means that the CAPM consists of three branches, and the feature of FAPD is divided into $3\times3$ regions, $4\times4$ regions, and $5\times5$ regions in these three branches respectively.}
  \vspace{3mm}
    \begin{tabular}{c|c|c}
    \hline
    \tabincell{c}{Method} & \tabincell{c}{WildSketch} & \tabincell{c}{CUFSF}  \\
    \hline\hline
     3 & 36.19 & 66.01\\
     \hline
     4 & 36.36 & 66.01 \\
     \hline
     5 & 36.21 & 66.14\\
     \hline
     3,4 & 36.86 & 66.36\\
     \hline
     4,5 & 36.84 & 66.40\\
    \hline
    3,4,5 & \bf{37.28} & \bf{66.84}\\
    \hline
    3,4,5,6 & 37.11 & 66.74\\
    \hline
    \end{tabular}
  \vspace{2mm}
  \label{tab:adaptive_ablation}
\end{table}

\subsubsection{Exploration of Component-Adaptive Perceiving Module}
In this subsection, we first verify the effectiveness of our CAPM. As shown in Table \ref{tab:ablation}, compared with ``FCE+FAPD", our model ``FCE+FAPD+CAPM'' achieves an improvement of Scoot 1.54\% on WildSketch dataset and 3.21\% on the CUFSF datasets. Furthermore, we analyze the input and output features of CAPM visually. As shown in Figure \ref{fig:adaptive_visual}, after applying the CAPM, on some regions that contain certain facial components (e.g., eye and mouth), their features are enhanced, which results in better sketch patches. These quantitative and qualitative experiments well demonstrate the effectiveness of the proposed CAPM.

We then explore different settings of CAPM by dividing the feature of FAPD into different numbers of regions. As shown in Table \ref{tab:adaptive_ablation}, when the feature is divided into single-scale regions (such as the $3\times3$ regions), the improvement is remarkable compared with ``FCE+FAPD", since some facial components are well modeled. When combining the features of multi-scale regions, the performance can be significantly improved, since the pyramid adaptive convolutions can well address the face misalignment problem. In particular, our model achieves a state-of-the-art Scoot of 37.28\% on WildSketch and 66.84\% on CUFSF, when the CAPM adopts a setting of ``3,4,5''. Under this setting, the CAPM consists of three branches, and the feature of FAPD is divided into $3\times3$ regions, $4\times4$ regions, and $5\times5$ regions in three branches respectively. When further combining the feature of $6\times6$ regions, the performance slightly drops, since the excessively small regions may fail to cover the whole facial components and hence leads to side effects. Therefore, our final CAPM adopts the setting of ``3,4,5''.

\begin{figure}[t]
\begin{center}
 \includegraphics[width=1\columnwidth]{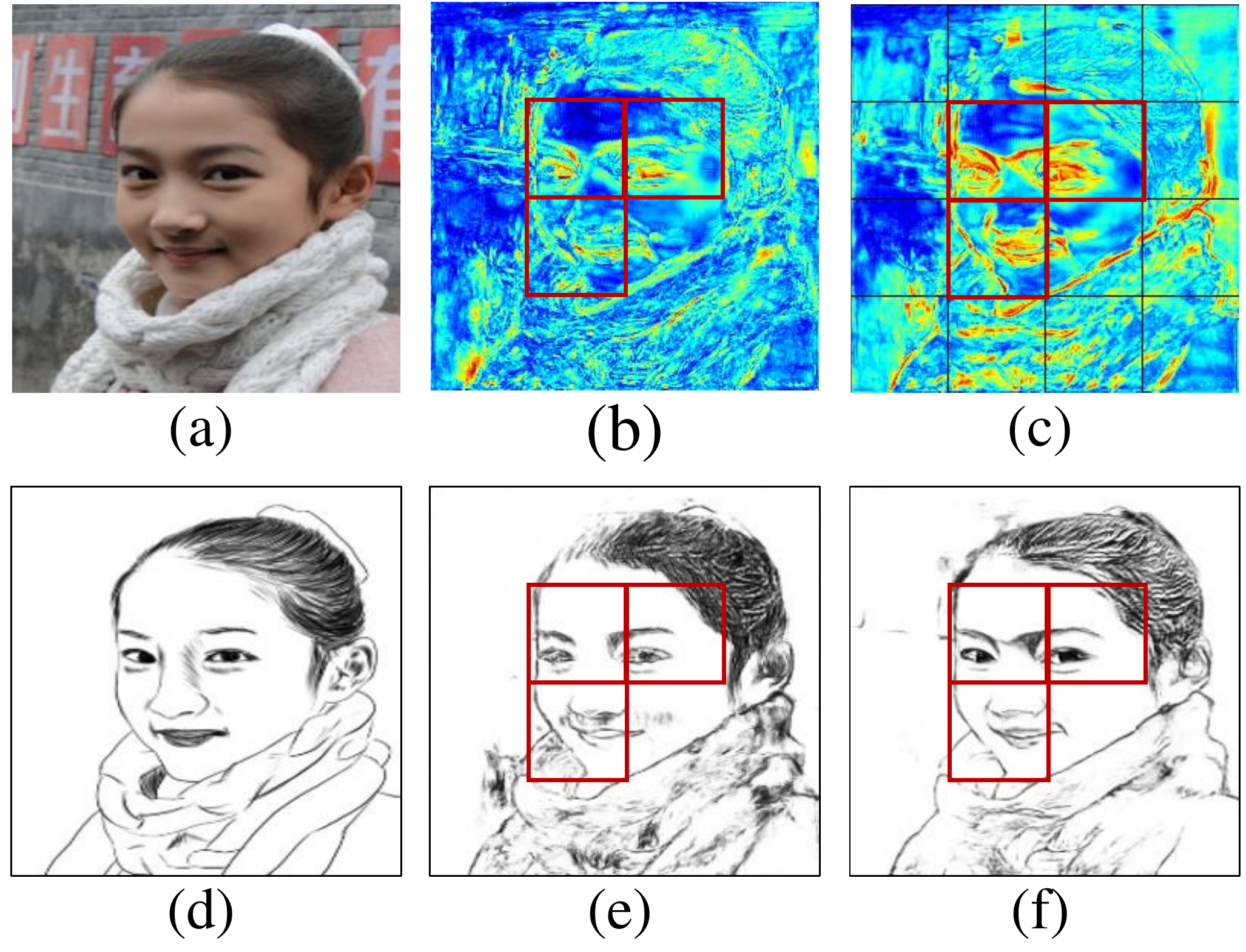}
 \vspace{-11mm}
\end{center}
   \caption{(a) An unconstrained image; (b) The output feature of FAPD; (c) The output feature of the second branch ($4\times4$ regions) of CAPM; (d) The ground-truth face sketch of (a); (e) The synthesized sketch based on (b); (f) The final sketch. Note that the visualized feature (b) and (c) are the channel-wise average of their original features. }
\vspace{0mm}
\label{fig:adaptive_visual}
\end{figure}

{\Revise{
\subsection{Limitation}
In this work, we find that the contours of many synthesized sketches are not sharp and some edges are intermittent under the unconstrained setting, as shown in Fig. \ref{fig:Unconstrained_Result}. This may be because edge details are gradually lost in the encoding stage. Therefore, some techniques of contour preservation are desired to improve the qualities of unconstrained sketches. In future work, we would address this problem from two aspects, including contour-aware representation learning and contour-aware loss function optimization. For the former, the contours of input images are first extracted and then incorporated to learn contour-preserving features at the decoding stage. For the latter, we would apply some contour losses \cite{chen2019contour} to supervise the sketch synthesis of each facial component.}}

\section{Conclusion}\label{sec:conclusion}
In this paper, we introduce a novel Perception-Adaptive Network (PANet) for face sketch synthesis in the wild. Specifically, our PANet incorporates deformable convolutions and adaptive convolutions to effectively perceive facial region and facial components, thereby directly generating high-quality sketches of the given unconstrained face images without any preprocessing. To promote further research and evaluation of this task, we also construct a new database, which contains 800 pairs of face photo-sketch with huge variations of facial appearance and background. Extensive experiments show the superior performance of our proposed method under both the constrained and unconstrained conditions.

\section*{CRediT authorship contribution statement}
{\textbf{Lin Nie}}: Conceptualization, Writing - original draft.
{\textbf{Lingbo Liu}}: Methodology, Writing - review \& editing.
{\textbf{Zhengtao Wu}}: Writing - review \& editing, Validation.
{\textbf{Wenxiong Kang}}: Writing -review \& editing.

\section*{Declaration of Competing Interest}
The authors declare that they have no known competing financial interests or personal relationships that could have appeared to influence the work reported in this paper.

\bibliography{reference}

\end{document}